\documentclass[11pt]{article}
\usepackage[utf8]{inputenc}
\usepackage{amsmath}
\usepackage{amssymb}
\usepackage{epsfig}
\usepackage[hyphens]{url}

\title{Quantifying Natural and Artificial Intelligence in Robots and Natural Systems with an Algorithmic Behavioural Test}
\author{Hector Zenil\\Unit of Computational Medicine, Stockholm, Sweden\\ \& Algorithmic Nature Group, LABoRES, Paris, France}
\date{}

\begin{document}

\maketitle

\begin{abstract}
One of the most important aims of the fields of robotics, artificial intelligence and artificial life is the design and construction of systems and machines as versatile and as reliable as living organisms at performing high level human-like tasks. But how are we to evaluate artificial systems if we are not certain how to measure these capacities in living systems, let alone how to define life or intelligence? Here I survey a concrete metric towards measuring abstract properties of natural and artificial systems, such as the ability to react to the environment and to control one's own behaviour.\\

\textbf{Keywords:} \textit{Natural computing; systems' behaviour; controllability; programmability; Turing test; compressibility; Kolmogorov complexity; randomness; robotics; artificial life}
\end{abstract}

\section{Introduction}

One key property of living systems is their sensitivity to external stimuli, which allows them to adapt and cope with the challenges of their complex environments. Indeed, a living organism will follow the complexity of its environment in order to survive~\cite{zenilentropy}. Engineered systems can be thought of as roughly emulating these properties of living systems, building programs and machines that react to external inputs, which are converted  into useful calculations or operations--abstract or physical--in a comparable fashion. 

One important question we face is how we are to evaluate artificial systems in a fashion comparable to natural systems, when it is so difficult to determine measures of performance for natural systems in the first place. For intelligence, for instance. One example of a test to measure it is the so-called Intellectual Quotient (IQ) test. But critics assert that it only measures how good a person is at taking IQ tests. Which brings to mind the fact that writing a general computer program to answer arbitrary questions on an IQ test is relatively easy once these questions are classified under the headings of arithmetic, word relationships and shapes, for example. We certainly do not mean to reduce human intelligence to something that can so easily be achieved by a relatively easy-to-code computer program. 

One may give up on absolutely objective tests of intelligence--just as Alan Turing himself did in order to make progress--conceding that the question of what constitutes intelligence, not only in humans but in machines, is difficult, if not impossible, to answer in a definitive way. And become reconciled to the fact that our only chance is to determine whether or not something looks as if it is performing a certain task in an intelligent fashion, regardless of the manner in which it goes about it. Thus a computer program passing a typical IQ test would be deemed just as intelligent as a human being passing the same test. Indeed, in biology for example, something is deemed alive if it manifests certain processes or activities attributed to life, such as growth, excretion and replication.

 This behavioural approach is also interesting because ultimately intelligence is an observer-dependent property. While something may appear intelligent to one person, it may not appear so to someone with a different set of expectations, either lower or higher. For example, machines are actually much better at many tasks than human beings--even classification tasks that we used to think humans were better at~\cite{ciresan}. They are better at playing games such as chess or Jeopardy! and answering factual questions---witness software such as IBM Watson and Wolfram Alpha. In fact, distinguishing a computer from a human is extremely easy, as computers can perform millions of sophisticated arithmetic operations in the time that it takes a human to perform just the simplest ones. Yet, Watson would fail terribly at arithmetic, just as an athlete would at designing a bridge, unless he also happened to be a civil engineer. Just as a submarine would fail terribly at flying or an airplane at swimming.

I find therefore, that it is ultimately necessary not only to admit the ineluctability of subjectivity but to seek natural measures that accommodate it, i.e., measures that are well equipped for context sensitivity. I think the measures I put forward are interesting in this regard, introducing the necessary objectivity while retaining the necessary subjectivity. Their theoretical basis is algorithmic information theory (denoted by AIT), with Kolmogorov complexity (which we will denote by $K$) as the central measure.

 Let's say one is presented with two strings. One is a sequence of one hundred 1s, the other is a random-looking string of length one-hundred bits. The Kolmogorov complexity of a string is defined as the length in bits of the smallest computer program that produces the string. So were you to be asked which of these strings looked more and which less random, according to traditional probability theory you would not be justified in saying that the string comprising only 1s is the less random one, because the two strings would have exactly the same probability of being produced by a uniformly random process generating strings of the same length. 

But using Kolmogorov complexity one can find a short program that produces non-random looking strings and justify their non-random character. For example, this program produces the string of 1s: $For\textnormal{ }n=1,\textnormal{ }While\textnormal{ }n=100,\textnormal{ }print(1),\textnormal{ }n = n + 1$. The program can produce a string of arbitrary length with the same pattern without having to increase the program length itself, except for changing $n=100$ to, for example, $n=1000$. But notice that while the string length increases to 1000 bits, the program length only increases by 1 bit, so the Kolmogorov complexity of a string of 1s grows by only a logarithmic factor. This means that this string of 1s is of the lowest possible algorithmic (or Kolmogorov) complexity. In other words, this program is the compressed version of an arbitrarily long string containing this pattern, and therefore the string is not considered (algorithmically) random. On the contrary, for a random looking string $s$ with no apparent pattern, the only way to produce a program for it would be a $print(s)$ which would be slightly longer than $s$ itself if $s$ cannot truly be compressed (which is our initial hypothesis). Therefore $s$ is said to be (algorithmically) random.

\section{Wolfram's classes of behaviour as a first case study}

\begin{figure}
\centering
\scalebox{.3}{\includegraphics{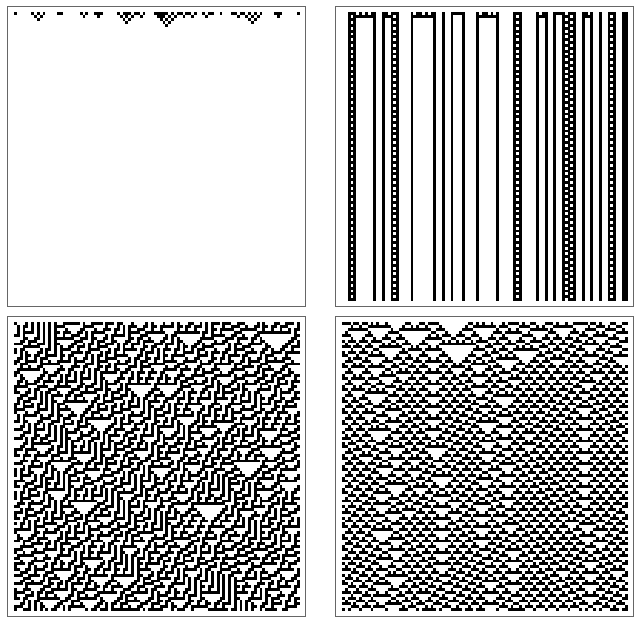}}
\caption{Archetypal cases of each of the 4 classes of behaviour identified by Stephen Wolfram in computer programs. Depicted here are four computer programs of so-called Elementary Cellular Automata (ECA) with rule numbers (from top left to bottom right) 160, 108, 30 and 54, starting from random initial conditions of length 100 with 0-1 density $\sim$ 0.5 running for 100 steps. Rule 30 is capable of what Wolfram calls \textit{intrinsic randomness}, even when starting from the simplest possible initial condition--a single black cell. Rule 54 has been conjectured to be Turing-complete~\cite{wolfram}, just like another class 4 ECA, Rule 110~\cite{wolfram,cook}, meaning that it is capable of simulating any other computer program.}
\label{ca}
\end{figure}

Before returning to Kolmogorov complexity let me introduce an experiment and a classification that Stephen Wolfram proposed, and that is relevant and interesting as a case study for our measures. Stephen Wolfram found~\cite{wolfram} that if one examined the space-time evolution diagrams of all computer programs, (see Fig.~\ref{ca}) one would find 4 types (Classes) of behaviour. He placed in Class 1 systems that evolve into homogeneous states, hence display simple behaviour. Class 2 systems develop into periodic states, such as fractals and crystal-like objects. Class 3 systems are random-looking, even if deterministic. And Class 4 systems display persistent structures over time, with some parts looking random and other parts looking simple. 

An elementary cellular automaton (ECA)~\cite{wolfram} is defined by a local function $f:\{0, 1\}^3 \rightarrow \{0, 1\}$. $f$ maps the state of a cell and its two immediate neighbours (range $= 1$) to a new cell state: $f_t:r_{-1},r_0,r_{+1} \rightarrow r_0$. Cells are updated synchronously according to $f$ over the space. 

\begin{figure}
\centering
\scalebox{.37}{\includegraphics{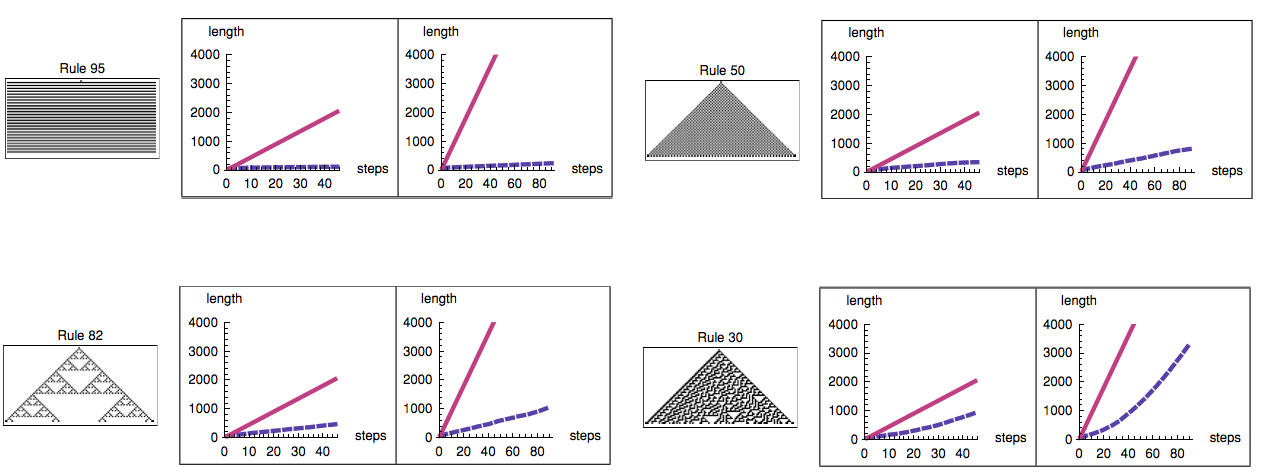}}
\caption{Identifying the behaviour of a system by looking at the asymptotic behaviour of the compressed lengths of its space-time evolutions. Plots show how more complex rules are increasingly hard to compress over time. This is a basic measure of complexity based on compressibility behaviour.}
\label{compressibility}
\end{figure}

One initial question of moment to try to address is whether Kolmogorov complexity, when applied to space-time diagrams, would be able to identify classes, assigning each a distinct measure. Fig.~\ref{compressibility} provides some examples of the evolution of cellular automata for specific numbers of steps and their compressed and uncompressed lengths using a lossless compression algorithm, that is, a compression algorithm that is able to recover the original object (cellular automaton in this case) when decompressing it. One can see that the evolution of class 3 cellular automata such as Rule 30 (top right in Fig.~\ref{ca} or bottom right in Fig.~\ref{compressibility}) are very hard to compress, while simple ones are very easy to compress. However, notice that if something is hard to compress with an algorithm, that doesn't mean it cannot be compressed. This is why the compressibility approach based on Kolmogorov complexity retains some subjectivity that I find useful in classifying systems. The subjectivity of Kolmogorov complexity is owed to the fact that the measure is semi-computable, which means that no Turing machine can compute the length of the shortest program for a given string. So only approximations, or upper bounds, to be more precise, are possible. But this is a desirable property of a measure of complexity, since it is because of its power that it turns out to be semi-computable, and its upper semicomputability means that compression is a sufficient test for non-randomness. That is, if something is compressible then we can be certain it has low Kolmogorov complexity, regardless of the problem of semi-computability.

Based on the principles of algorithmic complexity, one can characterise the behaviour of a system by comparing the result of the compression algorithms applied to its evolution to the system's uncompressed evolution~\cite{zenilchaos} . If the evolution is too random, the compressed version will not be much shorter than the length of the original evolution itself. It is clear that one can quantify this behaviour~\cite{zenilca}: if they are compressible they are simple, otherwise they are complex (random-looking). This approach can be taken further and used to detect phase transitions, as shown in~\cite{zenilca}, for one can detect differences between the compressed versions of the behaviour of a system for different initial configurations. This second measure allows us to characterise systems by their sensitivity to the environment: the more sensitive the greater the variation in length of the compressed evolutions. A classification places at the top systems that can be considered to be both efficient information carriers and highly programmable, given that they react succinctly to input perturbations. Systems that are too perturbable, however, do not show phase transitions and are grouped as inefficient information carriers. The efficiency requirement is to avoid what is known as Turing tarpits~\cite{perlis}, that is, systems that are capable of universal computation but are actually very hard to program. This means that there is a difference between what can be achieved in principle and the practical ability of a system to perform a task. This approach is therefore sensitive to the practicalities of programming a system rather than to its potential theoretical capability of being programmed.

\section{A subjective Turing test-like test for complex behaviour and stimuli sensitivity}

\begin{figure}
\centering
\scalebox{.3}{\includegraphics{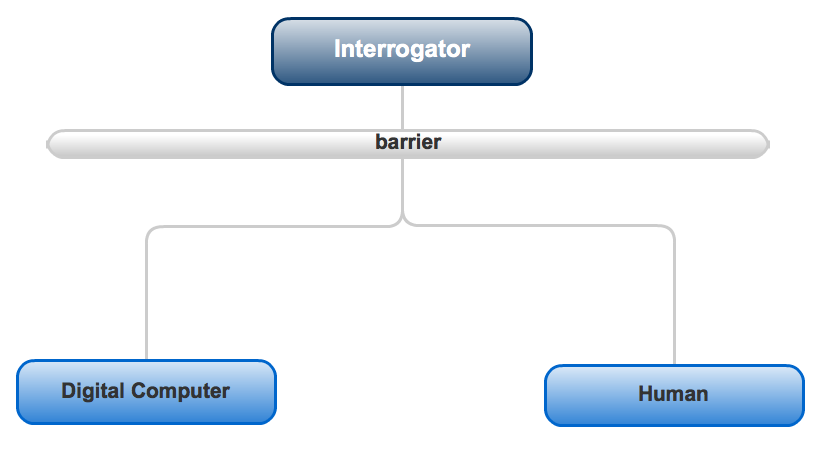}}
\caption{Original Turing test set as devised by Alan Turing where a computer aims at fooling an interrogator who has access to answers given by both a computer and a human but no way of knowing the source of the answers.}
\label{turingtest}
\end{figure}

This approach is very similar to another important contribution of Alan Turing to science, viz. Turing's imitation game, also known as the Turing test. The original Turing test consists in determining whether hidden behind a screen and answering an interrogator's questions is a human or a computer. Turing used this approach to give a possible answer to the question of whether computers could be intelligent, giving birth to the field of artificial intelligence. He replaced the question with a test, claiming that if a computer passed the Turing test, that is, if it succeeded in fooling a human frequently enough into believing it to be human, then the computer had to be regarded as intelligent (see Fig.~\ref{turingtest}). 

Here we are interested in the question of what kinds of measures can be applied to an artificial system, such as a robot or a chemical cell, or a natural system (such as slime mould), to determine whether it has some properties of living organisms, such as sensitivity, adaptability and controllability. I consider these properties to be properties of computation; sensitivity and adaptability are a form of input and output of a system, while controllability, a form of learning, can be compared to a computer program. At least this is the way we will explore some measures that can be used to assess properties of natural and artificial systems. In fact similar tests have been proposed to recognize artificial cells~\cite{krasnogor}, for artificial life~\cite{ecal}, artificial gene expression data~\cite{maier}, computation~\cite{zenilpt,brussels} and even in robotics~\cite{robots}, all having a similar motivation but a different approach as compared to the measures surveyed herein. 

Sensitivity tests in mathematical models of dynamical systems are common and well defined, but they are difficult to apply to systems where no equational form exists and where there is no well-defined metric distance on initial conditions. Here I present some methods and tools useful for tackling these questions from a pragmatic point of view. 

Sensitivity analysis is the study of how the uncertainty in the output of a mathematical model or system can be associated with different sources of uncertainty in its inputs. Sensitivity analysis can be useful for testing the robustness and variability of a system, and it sheds light on the  relationships between input and output in a system. This is not hard to identify from a computing perspective where inputs to computer programs produce an output, and computer programs can behave in different ways. 

Sensitivity measures aim to quantify this uncertainty and its propagation through a system. Among common ways to quantify and study this phenomenon is the so-called Lyapunov exponent approach. This approach consists in looking at the differences that arbitrarily close initial conditions produce in the output of a system. Traditionally, if the exponent is large the sensitivity is non-linear and divergence increases over time. If constant, however, the system is simple under this view.

 In our approach to computation the compression algorithm can be seen as an interrogator of the programming capabilities of a system, where the questions are initial conditions and the answer is the lengths of the compressed answers. If the source is somehow programmable then one should be able to declare that source able to compute. But unlike in Turing's test, I would like a measure of computation, a measure indicating whether I can actually program something to behave like a standard computer, and so I have come up with a measure based on compressibility that may be found in any of my papers on this subject.

\section{Towards a measure of programmability}
\label{main}

So if the measure of behaviour that I am proposing is based on lossless compression and ultimately Kolmogorov complexity, the measure will be highly dependent on how a system produces behaviour that looks or does not look (algorithmically) random. In general, for an intelligent robot, for example, one would wish to avoid the two extreme cases. A robot that does not show any signs of behaviour or reaction to external stimuli or that behaves in a random fashion should and will not be considered intelligent, regardless of the observer. There are also robots designed for the purpose of performing very repetitive tasks, whose behavior will actually be simple, albeit generative of structure. For a general purpose machine or system, however, a reasonable condition is to expect it to react to its environment, which is a sound requirement for intelligent behaviour, in the spirit of, for example, the Turing test itself. 

So the behavioural definition is based on whether a system is capable of reacting to the environment---the input---as reflected in a measure of \textit{programmability}. The measure quantifies the sensitivity of a system to external stimuli and will be used to define the susceptibility of a system to being (efficiently) programmed, in the context of a natural computing-like definition of behaviour~\cite{zenilpt,zeniljetai}.

Turing's observer is replaced by a lossless compression algorithm, which has subjective qualities just like a regular observer, in that it can only partially ``see" regularities in data, there being no perfectly effective compression algorithm in existence. The compression algorithm will look at the evolution of a system and determine, by means of feeding the system with different initial conditions (which is analogous to questioning it), whether it reacts to external stimuli.

 The compressed version of the evolution of a system is an approximation of its algorithmic (Kolmogorov) complexity defined by~\cite{kolmo,chaitin}:

\begin{center}
$K_T(s) = \min \{|p|, T(p)=s\}$
\end{center}

That is, the length of the shortest program $p$ that outputs the string $s$ running on a universal Turing machine $T$). A technical inconvenience of $K$ as a function taking $s$ to be the length of the shortest program that produces $s$ is its non-computability, proven by reduction to the halting problem. In other words, there is no program which takes a string $s$ as input and produces the integer $K(s)$ as output. This is usually taken to be a major problem, but one would expect a universal measure of complexity to have such a property. The measure was first conceived to define randomness and is today the accepted objective mathematical measure of complexity, among other reasons because it has been proven to be mathematically robust (in that it represents the convergence of several independent definitions). The mathematical theory of randomness has proven that properties of random objects can be captured by non-computable measures. One can, for example, approach $K$ using lossless compression algorithms that detect regularities in order to compress data. The value of the compressibility method is that the compression of a string as an approximation of $K$ is a sufficient test of non-randomness. If the shortest program producing $s$ is larger than $|s|$ the length of $s$, then $s$ is considered to be random.

A formal approximation to one such measure can be devised as follows. Let $C$ be an approximation to $K$ (given that $K$ is non-computable) by any means, for example, by using lossless compression algorithms or using the coding theorem technique we presented in \cite{delahayezenil,plosone}. Let's define the function $f$ as the variability of a system $M$ as the result of fitting a curve $\phi$ (by (linear) regression analysis) to the data points produced by different runs of increasing time $t^\prime$ (for fixed $n$) up to a given time $t$, of the sums of the differences in length of the approximations to Kolmogorov complexity ($C$) of a system $M$ for inputs $i_j$, $j\in\{1, \ldots, n\} \in E$, divided by $t(n-1)$ (for the sole purpose of \emph{normalising} the measure by the system's ``volume," so that one can roughly compare different systems for different $n$ and different $t$). With $E$ an enumeration of initial inputs for $M$. The following expression is a more formal attempt to capture the compressed lengths of $M$ for different initial conditions $i_j$. $M_t(i)$ is a system $M$ running for time $t$ and initial input configuration $i$. At the limit $\mathbb{C}_t^n$ captures the behaviour of $M_t$ for $t \rightarrow \infty$, but the value of $\mathbb{C}_t^n$ depends on the choices of $t$ and $n$ (we may sometimes refer to $\mathbb{C}$ as assuming a certain $t$ and $n$), so one can only aim to capture some average or asymptotic behaviour, if any (because no convergence is guaranteed). $\mathbb{C}_t^n$ is, however, an indicator of the degree of programmability of a system $M$ relative to its external stimuli (input $i$). The larger the derivative, the greater the variation in $M$, and hence in the possibility of programming $M$ to perform a task or transmit information at a rate captured by $\mathbb{C}_t^n$ itself (that is, whether for a small set of initial configurations $M$ produces a single significant change or does so incrementally).  Now the second step is to define the asymptotic measure, that is, the derivative of $f$ with respect to time, as a system's programmability (first basic definition):

\begin{equation}
\label{index}
\centering
\mathbb{C}_t^n(M)= \frac{\partial f(M,t,n)}{\partial t}
\end{equation}

For example, as is shown in \cite{zenilca}, certain elementary cellular automata rules that are highly sensitive to initial conditions and present phase transitions which dramatically change their qualitative behaviour when starting from different initial configurations can be characterised by these qualitative properties. A further investigation of the relation between this transition coefficient and the computational capabilities of certain known (Turing) universal machines has been undertaken in \cite{zeniluniversalca}. We will refrain from exact evaluations of $\mathbb{C}$ to avoid distracting the reader with numerical approximations that may detract from our particular goal in this paper. Other calculations have been advanced in \cite{zenilpt} and \cite{zeniljetai}.

In fact $\mathbb{C}$ is a family of measures, one for each lossless compression algorithm, hence mirroring the observer-dependent nature of the behaviour of one system relative to another.

\section{A Turing-test like test for computation and intelligent behavior}

The approach presented here can deal with situations such as Chalmer's rocks or Floridi's envelopes. For it permits one to assign a very low computing value to a rock, indicating that indeed, in agreement with our intuition, a rock is not like a computer because it is very difficult, if not impossible, to program a rock to carry out a computation. And a robot enveloped in a friendly environment will also have a very low level of intelligence because of its inability to react to external stimuli when faced with an unfriendly environment. And things like computers, such as rule 110, have a large computing value, as is consistent with our knowledge of this cellular automaton that is known to be capable of Turing universal computation despite its simplicity. Hence central to my measure of computing--and just as one might expect--is the notion of programmability, of being able to program something to perform a task. And the greater the number of possible tasks that a computer can be reprogrammed to perform, the greater the computing value of the object/system.

\begin{figure}
\centering
\scalebox{.3}{\includegraphics{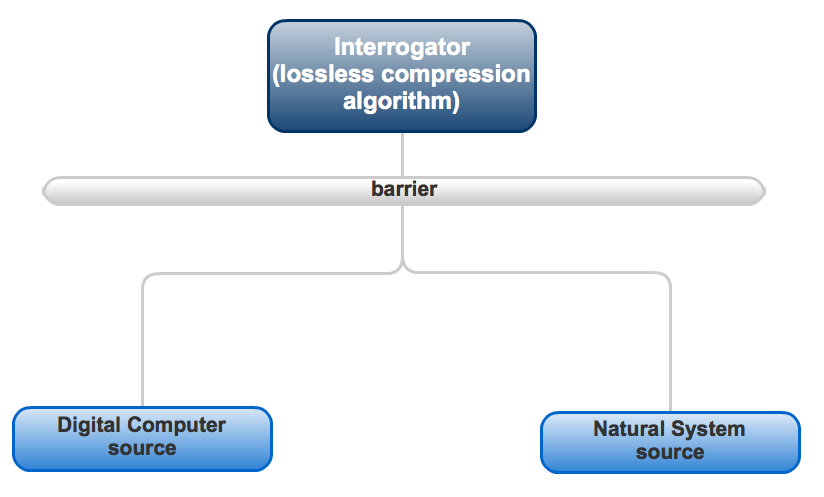}}
\caption{Modified Turing test to determine the behaviour of an artificial system in an automatic fashion using compression and to assess the complexity of the answers to questions in the form of random stimuli.}
\label{turingtest2}
\end{figure}

In Fig.~\ref{turingtest2}, a variation of the Turing test is proposed as a way to evaluate a system's ability to react to external stimuli--in an objective fashion while still being observer-dependent (the compression algorithm). One key question addressed in this paper is whether (algorithmic) information theory can be used to understand and quantify the behaviour of a natural or artificial system. For example, whether the spatial or morphological computation produced by a system can be understood and even manipulated by using tools drawn from (algorithmic) information theory.

\begin{figure}
\centering
\scalebox{.28}{\includegraphics{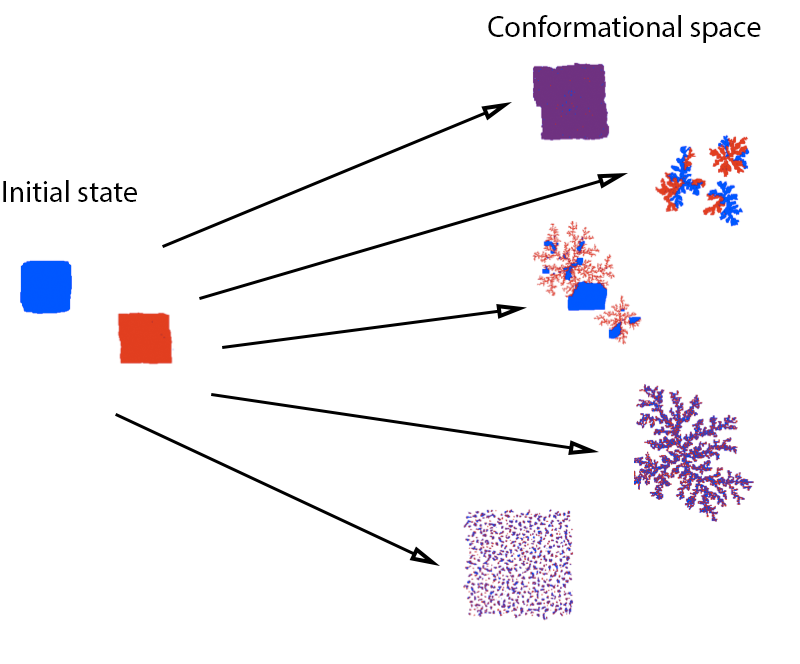}}
\caption{Mapping parameter space to conformation space of a system simulating the interaction of two types of biological (porphyrin) molecules with a Wang tile-based system. That is, a mapping of the input that makes the molecules self-organise into different shapes to perform different computational tasks.}
\label{conformationalspace}
\end{figure}

One of the aims of my research is to exploit all these ideas in order to try to reprogram living systems so as to make them do things as we would like them done, for in the end this is the whole idea behind programming something. Here is an example of finding the right input for porphyrin molecules to make them behave in different ways, that is, what it means to program something, to find the inputs for the desired output. It is all about mapping the parameter space to the output space, in this case the conformational space of these molecules, that is, the space of possible shapes. Fig.~\ref{conformationalspace} summarises an investigation recently undertaken by simulating the behavior of natural molecules called porphyrins that impart color to blood and help transport nutrients through the bloodstream.

 Indeed, in biology, a common challenge is the prediction of behaviour or shape, for example, examining RNA to predict how a protein will fold. Or predicting whether immune cells will differentiate in one fashion rather than another. And in Fig.~\ref{conformationalspace} what was investigated was how one could arrive at a certain specific conformational configuration from an initial state by changing environmental variables, such as temperature and other binding properties. The proofs that certain abstract systems implemented in software can reach Turing universality constitute one example of how hardware may be seen as software. Indeed, by taking simple four-colored tiles (called Wang Tiles, after Hao Wang) and placing them on a board according to the rule that whenever two tiles touch the touching sides must be of the same color, one can build an abstract machine that can simulate any other computer program. And this is a very powerful approach because one can program molecules or robots to do things like carry a payload to be released when certain specific conditions are met, releasing a chemical that may be used for biological markers, to fight disease or deal with nuclear garbage.

\section{A measure for robot behavioural complexity assessment}

According to a concept used in mechanical engineering and robotics a \textit{work envelope} (or \textit{reach envelope}) is a space representing all positions which may be occupied by an object during its normal range of motion. The work envelope hence defines the boundaries in which a robot can operate. One example is a dishwasher, where instead of a machine mimicking the way a human would wash dishes, there is an artificial friendly environment enveloping a very simple machine (e.g. a rotating stick) that gets the job done. In other words, we envelop simple robots in micro-environments that enable us to exploit their very limited capacities to deliver a desired output. A robotic arm's (see Fig.~\ref{arm}) envelope, if all joints are actuators with 3 degrees of freedom, is bounded by a sphere with centre the base of the arm and radius the length of the straight arm. The dishwasher clearly cannot be identified as intelligent except in its limited space. Floridi~\cite{floridieu} is right in that robotics as practised thus far proceeds largely in this way--by enveloping the world instead of embedding agents into it that are capable of reacting to and acting upon general environments, something which has traditionally been much harder to achieve. 

The measure surveyed herein is sensitive to Floridi's argument in the sense that highly enveloped technology is only intelligent in respect to its immediate environment, not to the larger (i.e. the outer) environment. In this sense, humans increase their intelligence to the degree to which they are in touch with their environment, but once separated from it (e.g. by computers) their capabilities become more restricted. This is all quantified by way of a mapping of parameters chosen in the domain of the environment. More formally, an agent $\alpha$ in an environment $E$ has intelligence $C(\alpha,E)$ but there exists an environment $E^\prime$ for which $C(\alpha,E) \neq C(\alpha,E^\prime)$. More specifically, one can in principle devise  environments $E^n$ such that the following chain forms a total order: $C(\alpha,E^{n-1}) \leq C(\alpha,E^{n}) \leq C(\alpha,E^{n+1})$ where $C$ is of course a measure in the family of measures defined in Section~\ref{main} based upon algorithmic complexity and approximated by lossless compression.

\begin{figure}
\centering
\scalebox{.37}{\includegraphics{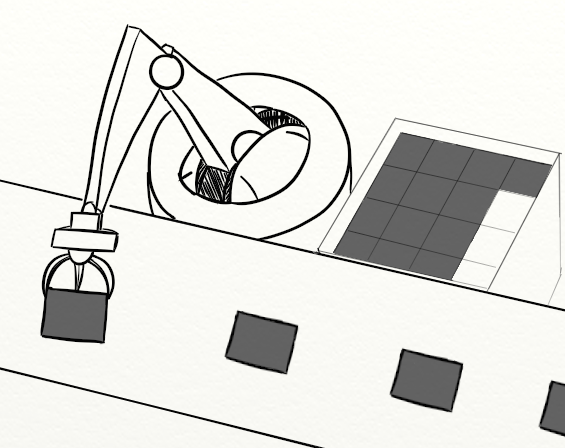}}
\caption{A robotic arm in an `assembly line'-type task engaged in a repetitive job is assigned low complexity if unable to react to external stimuli. Drawing by EVZ with love and kind permission to use.}
\label{arm}
\end{figure}

 A key concept that we have developed along these lines is that of spatial complexity. In~\cite{plosone} we introduced a measure defined naturally in two dimensions that allows direct application of a formal complexity measure to the evolution of systems, such as space-time diagrams, that would not be difficult to imagine applied to modular robots research~\cite{modularrobots}. The basic idea, which is also related to the main measure~\ref{main}, is the characterisation of movement in 2-dimensional projections, so that one can then determine the complexity of a lattice where black cells represent the trajectory of a machine's movement by determining the probability of a Turing machine following a random program reproducing the same pattern. The empty lattice, for example, is the one with lowest complexity because a large fraction of random computer programs running on a (prefix-free) Turing machine will produce no output. Therefore it is assigned low complexity according to the relation established by the so-called algorithmic Coding theorem~\cite{levin} $K(x)=-\log p(x)$, where $K$ is the algorithmic complexity of an object $s$ and $p(x)$ is the probability of $x$ occurring as the output of a random computer program.

\begin{figure}
\centering
\scalebox{.31}{\includegraphics{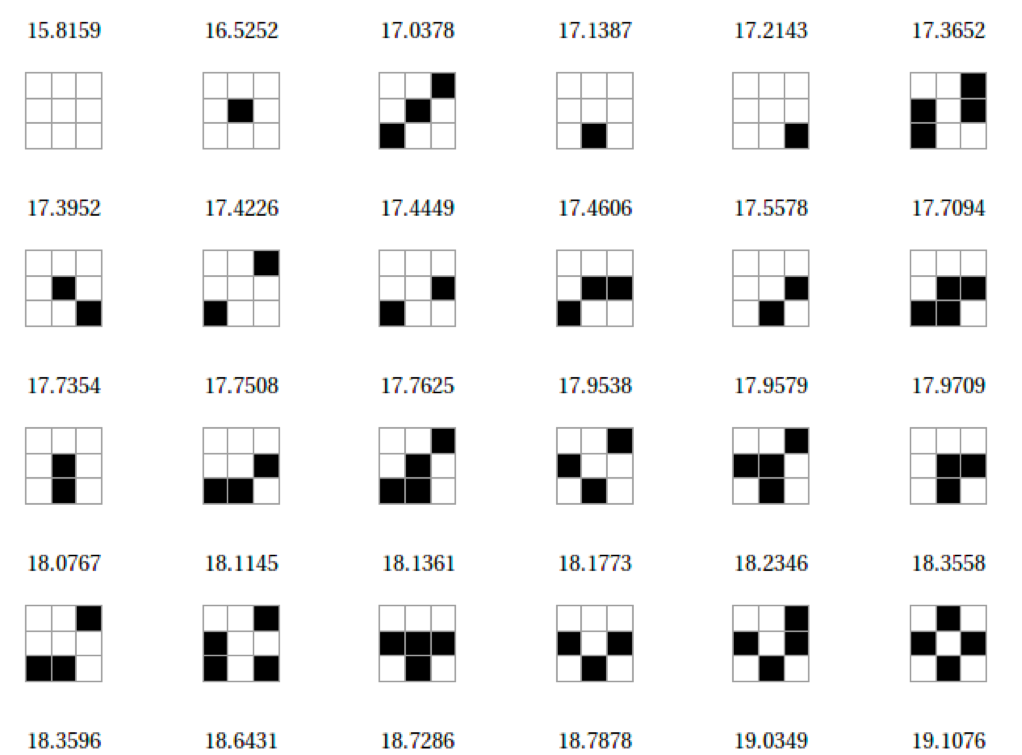}}
\caption{Simplified sketch of movements in a robotic work envelope after a simulated projection of a robot movement in a very coarse-grained small lattice. The more complex a robot's movement the higher the 2-dimensional Kolmogorov complexity.}
\label{arm}
\end{figure}

The measures presented can have other applications than in the area of computers. e.g. in measuring the programmability of physical and biological entities. Here is a sketch of how this would work, sorting various systems by variability and controllability, which are the two orthogonal (though not necessarily independent) properties that I think are fundamental to take into account in devising a measure for assessing a system's (or machine's) behavioural capabilities (see Fig.~\ref{programmability}). Additionally, one can easily expand the scope of the measure in order to have it serve as a relative or conditional measure for comparing the behaviour of two systems, for example, one tested and one target system. An instance would be a robotic hand capable of holding fragile objects versus a human hand. The robotic hand will have sensors to evaluate the weight and fragility of the object to enable it to decide how and with how much pressure to hold it. If the behavioural approach yields a result similar to a representation of the hand's behavioural output for this task, then the robot should be deemed successful. It is clear that on many occasions the choice of behavioural representation will be particularly difficult. For this example, one can devise an adequate 2-dimensional projection of the robotic and human hands (the same for both) associated with whether or not the object has been successfully held. One will then have a list of associations of successful movements and parameters (e.g. weight, infringed pressure) to store in digital form for behavioural evaluation.

 For example, let's take the example of a robotic arm (see Fig.~\ref{arm}). One way to represent its output behaviour is by projecting its movements on 3 orthogonal planes. One can see that this can be generalised to any robot whose action is to be evaluated, so in a 3-dimensional space 3 orthogonal planes will capture all movements. Then each of the coordinates in each plane, 6 coordinates for every state, will be coupled as a point in a 6-dimensional phase space. If the robotic arm is performing a repetitive task, as is the order of the day in car factories, the phase space will show a periodic pattern with low algorithmic complexity approximated by high lossless compressibility. The next step is to pair the input stimuli to the output for every time step. For a robotic arm in a car factory with no reaction to external stimuli, the input parameter space will lead to no behavioural reaction and therefore the arm will be considered to be of very low behavioural capability (or \textit{intelligence}), as the only way to reprogram it is within the system itself, i.e. by opening up the arm or writing a new computer program to control the arm and hardcode any new behaviour while the arm is disconnected from the environment or indirectly connected to it by a third system (e.g. an external observer---the programmer). However, if the arm reacts to external stimuli, the correspondence between parameter and behavioural space will start showing more complicated patterns, producing correlations between inputs and outputs, hence exhibiting some sort of adaptability and sensitivity to the environment. It will therefore be assigned a higher algorithmic complexity by virtue of the lower compressibility of its phase space.

\section{Conclusions}

\begin{figure}
\centering
\scalebox{.32}{\includegraphics{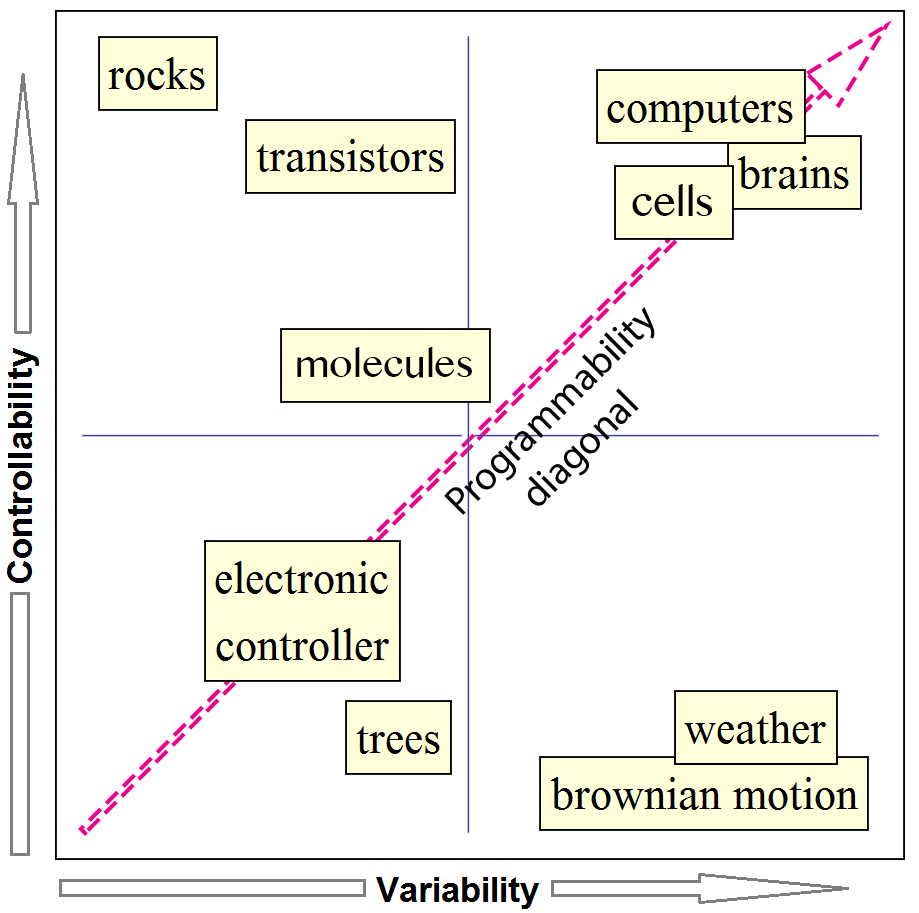}}
\caption{The behavioural measures here presented and surveyed suggest a natural classification of the programmability of natural and artificial systems.}
\label{programmability}
\end{figure}

We have seen how similar natural and artificial machines and systems can be evaluated with a concept of behavioural programmability, both objective and quantitative, and subjective in nature, this latter property in agreement with some desirable properties for evaluation, such as how intelligent a system is with respect to another, or with respect to a specific expectation. The measure has been applied to the design and control of self-organized molecular computing using porphyrins~\cite{terrazas} and in psychology for subjective randomness quantification~\cite{gauvrit}. The family of measures based upon algorithmic complexity introduced and surveyed here are behavioural in nature, and similar in spirit to the Turing test for machine intelligence, but extending it in a precise fashion, providing grades by quantifying the reaction sensitivity of a system to external stimuli from the observer perspective. They may potentially be applied to machine performance estimation, where adaptability and sensitivity to the environment is key, and where the availability of resources must be taken into consideration. 

The behavioural approach in fact generates a natural classification of objects by their programmability, as sketched in Fig.~\ref{programmability}. For example, while weather phenomena and Brownian motion have great variability, they are hardly controllable. On the other hand, rocks have a very low variability and are therefore trivially controllable but are not therefore on the programmability diagonal and cannot count as computers. Everything on the diagonal, however, including living organisms, is programmable to some extent. In this way I think the question of measuring intelligent properties of natural and artificial systems can be approached.

\subsection*{Acknowledgement}

H. Zenil thanks the Foundational Questions Institute (FQXi) for the grant awarded on ``New directions in research on cellular automata and information biology'' (FQXi-MGA-1414).

\end{document}